\def\year{2022}\relax
\documentclass[letterpaper]{article} 
\usepackage{aaai22}  
\usepackage{times}  
\usepackage{helvet}  
\usepackage{courier}  
\usepackage[hyphens]{url}  
\usepackage{graphicx} 
\usepackage{natbib}  
\usepackage{caption} 
\DeclareCaptionStyle{ruled}{labelfont=normalfont,labelsep=colon,strut=off} 
\usepackage{algorithm}
\usepackage{algorithmic}
\usepackage{enumitem}
\usepackage{amsmath}
\usepackage{amsfonts}

\usepackage{newfloat}
\usepackage{listings}
\floatstyle{ruled}
\newfloat{listing}{tb}{lst}{}
\floatname{listing}{Listing}
\title{Incentives in Two-sided Matching Markets with Prediction-enhanced Preference-formation}

\author{
    Stefania Ionescu$^1$,
    Yuhao Du$^2$,
    Kenneth Joseph$^2$,
    Anikó Hannák$^1$
}
\affiliations{

    $^1$ University of Zürich, 
    $^2$ University at Buffalo\\
}

\nocopyright

\begin{document}

\maketitle

\begin{abstract}
Two-sided matching markets have long existed to pair agents in the absence of regulated exchanges.  A common example is school choice, where a \emph{matching} mechanism uses student and school preferences to assign students to schools. In such settings, forming preferences is both difficult and critical. Prior work has suggested various \emph{prediction} mechanisms that help agents make decisions about their preferences. Although often deployed together, these matching and prediction mechanisms are almost always analyzed separately. The present work shows that at the intersection of the two lies a previously unexplored type of strategic behavior: agents returning to the market (e.g., schools) can attack future predictions by interacting short-term non-optimally with their matches. Here, we first introduce this type of strategic behavior, which we call an \emph{adversarial interaction attack}. Next, we construct a formal economic model that captures the feedback loop between prediction mechanisms designed to assist agents and the matching mechanism used to pair them. This economic model allows us to analyze adversarial interaction attacks. Finally, using school choice as an example, we build a simulation to show that, as the trust in and accuracy of predictions increases, schools gain progressively more by initiating an adversarial interaction attack. We also show that this attack increases inequality in the student population. 
\end{abstract}


\maketitle

\section{Introduction}
In \textit{two-sided matching markets}, agents are partitioned in two disjoint sets (e.g., students and schools) and want to get paired with agents from the other set for future bilateral exchanges (e.g., a student learns at a school and the school provides instruction to the student) \cite{roth1992two}. In such markets, 
agents typically report their preferences over potential matches and a \textit{matching mechanism} uses those preferences to produce an \textit{assignment}, i.e. a pairing between agents.
Because forming one's preference is hard, agents benefit from external recommendations.


Since predictive models have become increasingly accessible, reliable, and trusted \cite{logg2019algorithm}, they also became more frequently used to inform preference-formation in matching markets. 
One such example is school choice. 
After the introduction of the No Child Left Behind Act, in 2001, students from low-performing schools were allowed to transfer to better-performing schools. However, initially, only a few students took advantage of this opportunity, partly because it was hard for parents to assess which school would improve their child's performance. In 2008, a content-based recommender system called SmartChoice was deployed for focus group participants; its goal was to help parents with the assessment by identifying the best schools based on predictions on the student's development at that school \cite{wilson2009smartchoice}. 
Another example is in refugee assignment, where refugees are matched with locations. Here, the preferences of locations over refugees are given by a machine learning (ML) model that predicts the level of integration success (e.g., measured by the probability of employment within 90 days) of an individual at a given location \cite{bansak2018improving}. `The Swiss government has recently implemented a randomized test to examine the performance of data-driven algorithms for outcome-based assignment' \cite{ acharya2019combining}.
Similar approaches have been developed for other application domains such as labor market \cite{paparrizos2011machine} and course allocation \cite{kurniadi2019proposed}.


These examples show that using predictive models to inform the preferences of agents in matching markets is not a problem of the future, but rather one of the present.
However, prior work has evaluated potential vulnerabilities \emph{separately} in the matching mechanisms \cite{erdil2008s, budish2012multi, abdulkadirouglu2003school} and  prediction-based recommendations \cite{o2004collaborative, mobasher2007toward}. When considering the matching algorithm, one usually analyzes the incentives of individuals by, for example, asking whether the mechanism is \textit{strategy-proof} (i.e. whether agents have an incentive to misreport their preferences). 
Similarly, there is a broad literature on  attacks on recommendations \cite{christakopoulou2019adversarial, o2004collaborative, mobasher2007toward} and predictive models which might face \textit{poisoning} or \textit{evasion} attacks (i.e. attacks which either inject fake data to trigger an unfaithful model or perturb the testing input to trigger a misclassification) \cite{huang2011adversarial}. 

In this paper, we argue that, in addition to considering vulnerabilities of the matching and prediction mechanisms independently, it is critical to also look at vulnerabilities in systems that \emph{combine them together}.
Specifically, we consider systems similar to those described above, i.e, where: (a) agents of one side (the \textit{returning side}) come back to the matching market in subsequent rounds (e.g., schools), (b) agents have post-matching objectives (e.g., schools want to be prestigious and minimize their cost), (c) agents on the returning side \emph{have the power to shape the quality of the interaction} with those they are assigned to via post-matching decisions (e.g,  schools can increase the performance of students through extra-curricular preparations or integration programs), and (d) these interactions impact outcomes for the non-returning side, which in turn influence future predictions (e.g., predictions based on the outcomes of past students change the preferences of current students). 

By making the feedback loop between the matching and prediction mechanisms explicit, we uncover a new type of strategic behavior:
the returning side can attack the system by changing their interactions with their matches. Even though in the short-run it might be beneficial for both matched parties to have the best possible interaction (e.g., both locations and refugees want refugees to be employed as soon as possible), the returning side (e.g., locations) might want to sacrifice their utility in the current round for different future predictions (e.g., locations might postpone hiring difficult-to-integrate refugees so they will not be allocated similar refugees in the future). 
We call such post-matching strategic behavior---where agents sacrifice their short-term utility to trigger different long-term predictions--- \textit{adversarial interaction attacks}. Similarly, we call a system where agents cannot benefit from such attacks \textit{interaction-proof}.

In this paper, we:
\begin{itemize}[nolistsep]
    \item Build a formal model for repeated two-sided matching markets with prediction-enhanced preference formation;
    \item Use this model to define \textit{adversarial interaction attacks} and the optimization problem faced by returning agents;
    \item Use an agent-based model to analyze when and by how much agents can benefit from interaction-strategic behavior. We show that, for some systems, (a) the returning agents have an incentive to use adversarial interaction attacks, (b) the utility gains obtained through such attacks increase as predictions become more accurate and trusted, (c) once a returning agent attacks, others have an incentive to implement more severe attacks, and (d) the non-returning agents are unevenly negatively affected by such attacks.
\end{itemize}


\section{Other Related Work}
\label{sec_RW}


\emph{Adversarial Attacks in Recommendations.}
From past literature in recommender systems (RS), the most similar adversarial attacks to the one we consider are \textit{shelling attacks} (i.e., attacks where fake users and ratings are created to trigger different future recommendations). These works mainly focused on crafting adversarial examples \cite{o2004collaborative, mobasher2007toward}. Differently, \citet{christakopoulou2019adversarial} analyzed shelling attacks from a machine learning and optimization perspective by building on the literature on poisoning attacks in classification tasks \cite{li2016data}, and formulating the problem as a two-player general-sum game between an RS and an Adversarial Attacker. 

However, our setting has some key differences. First, the returning agents usually cannot inject fake users (e.g., because there are public records of which students attended which schools). Instead, agents attack by adapting their actions. This distinction has both social implications (the non-returning agents experience a different outcome while the attack takes place) and economic implications (the returning agent implementing the attack sacrifices their short-term utility). Second, the existence of capacity constraints entails a rivalry for available seats (if a student is accepted at a school then another student is not accepted, and vice versa). Consequently, when attacking for being matched to a specific (type of) agent, there are two intents to do so: (a) rank higher in the preferences of agents who are of interest, and \emph{(b) rank lower in the preferences of agents who are not of interest}. The latter is different from classical recommender system applications and is increasingly important when the returning agents cannot fully and freely express their preferences. Third, also because of competition, agents do not decide to whom they are allocated (not all students can get a place to their most preferred school); instead, they report their preferences to a matching market, which produces an assignment. Therefore, attacks in this setting are not only efficient if they change the most preferred option of a non-returning agent, but also if they change the ordering of options lower in the preference ranking.

\emph{Strategic Behavior in Matching Markets.} 
As noted above, matching markets (MM) are widely used to pair agents based on their reported preferences; in such settings agents can behave strategically by, e.g., misreporting their preferences. Prior work shows that users find and implement such profitable manipulations which leads to congestion and inefficiencies \cite{budish2012multi}. Thus, strategy-proofness is a common desiderata in real-world application domains; e.g., in the Boston school choice system, this triggered a transition from the Boston to the Deferred Acceptance mechanism \cite{abdulkadiroglu2006changing}. 
In this paper we analyse strategic-interacting, which differs from strategic-reporting as it is not an attack on the mechanism alone, but rather on its combination with the prediction-based preference-formation process. 
When proposing predictive modelling for refugee assignment, \citet{bansak2018improving} distinguished between the modeling phase (when the predictive model is built) and the matching phase (when the assignment is produced). We generalise and extend this framework by also considering the phase when the paired agents interact and new data is produced. Moreover, we build on prior literature in economics when creating the model (for formalising the matching phase \cite{roth1982economics}) and when analyzing the system (for unilateral deviations and Nash equilibria \cite{myerson2013game}). 

\emph{Using Simulations to Understand Long-Term Effects.}
The machine learning community has seen an increasing use of simulation to study the interaction between users and recommendations. RecSim \cite{ie2019recsim} provides an environment that naturally supports sequential interaction with users. \citet{Masoud2020Feedback} proposed a method for simulating interactions between users and RSs to study the impact of the resulting feedback loop on the popularity bias amplification. \citet{Bountouridis2019SIREN} built a framework called SIREN to study how RSs will impact users' news consumption preference in the long term. Similarly, scholarship within MM used simulations to understand the effect of different design choices. \citet{erdil2008s} developed an experiment in order to compare the performance of two alternative matching algorithms in the school choice setting. In the context of online dating, \citet{ionescu2021agent} proposed an agent-based model (ABM) to test the effects of different platform interventions in reducing racial homogamy. At the intersection of ML and two-sided markets, \citet{malgonde2020taming} built an ABM to test the effects of introducing a two-sided RS within a complex adaptive business system. Similar to this previous work, we use simulations in the context of recommendation by developing an ABM. We, however, have a different goal: to understand how the characteristics of the market affect the incentives to use and effects of using adversarial interaction attacks.


\section{Problem Formulation}
\label{sec_PF}
\begin{figure}[t]
\centering
  \includegraphics[width=1\linewidth]{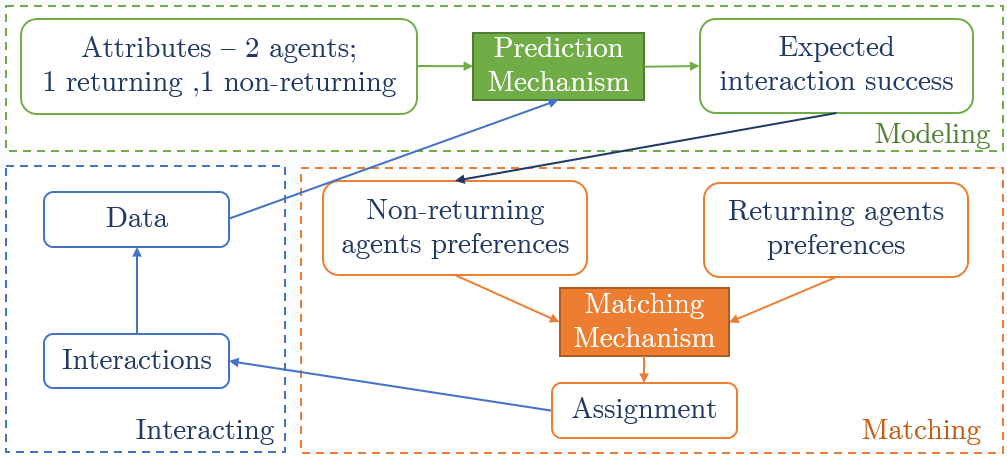}

  \caption{Overview of the three-phased system.}
\label{fig:model}
\end{figure}

Our model for the system is composed of three phases: (a) modeling, (b) matching, and (c) interacting. The modeling stage builds a predictive model that forecasts the interaction outcome of two agents if matched. During the matching phase, the non-returning agents use this model to inform their preferences
\footnote{Alternatively, the predictive model could be used to inform the preferences of the returning agents. For simplicity, we chose not to extend our model to capture this alternative in the main text and only include it in the appendix. Moreover, here we use school choice as a running example; see the appendix for other examples.}
; next, a \emph{matching algorithm} pairs each agent in the non-returning side of the market (student) with an agent in the returning side of the market.
Finally, in the interacting stage, the agents -- already paired according to the assignment obtained during matching -- interact with each other (students attend the classes at their assigned school). This interaction will produce an outcome (e.g., SAT scores of students) that is kept as a record and used to inform future predictions.
The entire system then repeats in a series of rounds with new non-returning agents each time, but with the same returning agents.
Figure~\ref{fig:model} shows an overview of all the phases and the interactions between them. In the remainder of this section, we formalise each of the stages and present the decision problem faced by the returning agents.

\subsection{The Matching Stage}
We start with the notation for the agents. Let $X^t$ be the set of non-returning agents to be matched at time (round) $t$, and $Y$ be the set of returning agents. Moreover, we denote by $\succ_x$ the preference of the non-returning agent $x\in X$ over the returning agents, $Y$. In other words, $\succ_x$ is a ranking over $Y$. Similarly, $\succ_y$ is the preference of $y\in Y$ over $X$. 
These preferences could be either unweighted (i.e., the agent only knows the ordering of their options) or weighted (i.e., the agent also knows how much more they prefer each potential match over another). 
In the weighted case, the preference $\succ_x$ has an associated weight function $w_x: Y\to \mathbb{R}$ mapping each option $y$ to the strength $x$ wants to be matched to $y$. 

A matching is a pairing of non-returning agents to returning ones, i.e., $\mu: X\to Y\cup \{\bot\}$. Here, $\mu(x) = \bot$ signifies that $x$ remained unassigned. More generally, a matching procedure (or allocation rule) maps the preferences of agents to a matching. In other words, a matching procedure $M$ is a function mapping every $(\{\succ_x\}_{x\in X}, \{\succ_y\}_{y\in Y})$ to a matching $\mu$. We denote by $\mu^t$ the matching in round $t$, i.e. $\mu^t = M((\{\succ_x\}_{x\in X^t}, \{\succ_y\}_{y\in Y}))$.


\subsection{The Interacting Stage}
After being matched, the agents interact. For every agent pair $(x, y)$, there is a set of possible outcomes, depending on their interaction. Since $x$ is not returning to the market, we assume they will always prefer the best outcome for them. Therefore, the set of possible outcomes depends on the actions of the returning agent, $y$. We denote the set of outcomes $y$ could choose from when interacting with $x$ by $\mathcal{O}_y(x)$.

Depending on the resulting outcome, each agent has some value, cost, and utility. If $y$ chooses outcome $o$ when interacting with $x$ we denote by $v_y^o(x)$ their value, by $c_y^o(x)$ their cost, and by $u^o_y(x) = v_y^o(x) - c_y^o(x)$ their utility. We use the analogous notation for $x$. We also make the simplifying assumptions that  (a) the value is symmetric, i.e. $v_y^o(x) = v_x^o(y)$, and (b) the cost of the non-returning side is null, i.e. that $c_x^o(y) = 0$ and $u^o_x(y) = v_y^o(x)$.

As an example, a school $y$ might have the choice set $\mathcal{O}_y(x) = \{80, 100\}$ since it could either invest in extracurricular preparation for $x$ (case in which $x$ scores around $100\%$ on state-administered tests) or not to invest (and $x$ scores around $80\%$). If school $y$ incurs a fixed cost of $5$ when preparing a student, then for the outcome $80$ both the student and the school have a utility of $80$ while for the outcome $100$ the student has a utility of $100$ while the school subtracts the cost from the value, and thus has a utility of $100-5 = 95$.

\subsection{The Modeling Stage}
The prediction model at time $t$ is informed by the history of interactions until that time, i.e.,
$H^t = \{(x, y, o, u)|y = \mu^u(x), y \text{ interacted with } x \text{ at time } u<t \text{ with outcome } o\in \mathcal{O}_y(x)\}$. The history thus records the agents matched so far and the time and outcome of their interaction.

The prediction algorithm uses the history as an input. Its output is a \emph{hypothesis}, i.e., a function which maps a pair of agents to their expected interaction outcome. We use the usual statistical framework in learning theory and denote the hypothesis by a parameterized function $h_{\theta_t}: X\times Y \to \mathbb{R}$. The prediction is thus based on the parameter $\theta_t$ at time $t$, which is obtained by solving the optimization problem $\theta_t = \arg\min_{\theta} L(\theta, H^{t-1})$. Here, $L$ is the loss function.

The hypothesis predicts values for a potential interaction between a non-returning agent, $x$, and all the returning agents. Therefore, it induces a preference weight function and a ranking over the returning agents. Formally, $h_{\theta_t}(x, \cdot): Y\to \mathbb{R}$ gives a weighted preference over $Y$, which we denote by $\succ^{h}_x$. By exposing agent $x$ to this ranking, their final preference might change. To capture this change, each agent $x$ has a \textit{prediction integration function}, $h_{\theta_t}^x$, that transforms a (weighted) preference of $x$ into a new preference, $\succ'_x$, using the hypothesis-based ranking, $\succ^{h}_x$.

To continue the previous example, let us assume that student $x$ initially believes schools $A$ and $B$ are equally good for them, i.e. $\succ_x$ is $A \sim B$ with, say, a weight function $w_x(A) = w_x(B) = 80$. Assume the hypothesis predicts $h_{\theta_t}(x, A) = 100$ and $h_{\theta_t}(x, B) = 80$, which corresponds to the ranking $A \succ^{h}_x B$. The exposure of student $x$ to this prediction-based ranking changes their preference to a new one which lies between their original opinion, $\succ_x$, and the suggested one, $\succ^{h}_x$. For instance, the new preference could correspond to the weight profile obtained by averaging the original and the suggested ones. Then, the new preference, $h_{\theta_t}^x(\succ_x) = \succ'_x$, has the weight function $w'_x(A) = 90$ and  $w'_x(B) = 80$ thus ranking $ A \succ'_x B$.

\subsection{The Decision Problem of Returning Agents}

The decision problem faced by the returning agents when they choose how to interact with their matches can be formalised as a dynamic programming task over the infinite horizon. They choose the action that maximizes the sum of their utility now plus the expected discounted utility in the future if they take that action. 
Using the notation introduced in the previous section, the maximum expected-discounted utility of a returning agent, $y$, from time $t$ onward is
\footnote{To achieve this succinct problem formulation, we made some simplifying assumptions and notations (details in the appendix).}
\begin{equation*} 
\begin{split}
U_y\left( \mu^t \right) = &\ max_{o\in \mathcal{O}_y(\mu^t(y))} \left( 
u_y^o\left( \mu^t(y) \right) + \right.\\
& \left. \beta \mathbb{E} \left[ U_y\left( \mu^{t+1} \right) | (\mu^t(y), y, o, t) \in H^{t+1} \right] \right),
\end{split}
\end{equation*}
where we make the standard assumption in economics that agents steeply discount future utility and denote the discount factor by $\beta$. The \textit{optimal strategy} of a returning agent is to take, at every round, the interaction leading to the outcome that maximizes the expected-discounted utility. 

The straightforward interaction strategy is to always choose the outcome maximising the 1-step utility, i.e. choose $\arg max_{o\in \mathcal{O}_y(\mu^t(y))} \left(u_y^o\left( \mu^t(y) \right)\right)$. When a returning agent uses this strategy, we say it \textit{interacts truthfully}. Naturally, in general, the optimal strategy needs not be the truthful one. The resulting gap leaves space for strategic interactions. We refer to strategies involving non-truthful interactions as \textit{adversarial interaction attacks} and call the systems in which agents cannot benefit from such attacks \textit{interaction-proof}.


\section{Experiment}
\label{sec_VE}
 We operationalize the framework above by creating and simulating an agent-based model (ABM) for school choice.  The parameters used in the experiment are summarised in Table~\ref{tab:ve}, and the code is available on GitHub \footnote{GitHub link: https://github.com/StefaniaI/Predictions-MM.}.


\begin{table}[t]
\footnotesize
\begin{tabular}{p{4.5 cm} p {3 cm}}
\hline
Parameters & Values Taken \\
\hline
{\bf Varied in Experiment}  & \\
\ \ \# of schools & 2, \textbf{10}\\
\ \ Attributes of students $\sim \mathcal{N}(\mu, \sigma)$ & $\mu \in \{\textbf{1}, 3.2\},$ \\
\ \  & $\sigma \in \{0.65, 1.5, \textbf{3}\}$ \\
\ \ Competition (\# students per place) & 4, \textbf{1}, 1/4\\
\ \ Sign of utility per student & \textbf{positive}, negative\\
\ \ Cost per improvement, $\alpha$ & 0.5, \textbf{0.95}\\
\ \ Matching mechanism & SD, \textbf{RSD}, Boston, DA\\
\ \ Level of prediction noise & \textbf{0.01\%}, 1\%, 10\%, 30\%\\ 
\ \ \# of past rounds used for training & 1, \textbf{3}, 5\\
\ \ \#  of neighbours (k) for KNN & \textbf{1}, 3, 5\\
\ \ Level of trust in recommendations & 0.5, \textbf{1}\\
\ \ Level of adversarial attack (in \%) & 0, \textbf{4}, 25, 50, 75, 100\\
\hline
{\bf Fixed in Experiment} & \\
\ \ \# of evaluation-relevant attributes  &  1\\
\ \ Evaluation scale for each attribute &  0 - 5\\
\ \ \# of students & $20\times \text{(\# of schools)}$\\
\ \ School capacity & $20 \div \text{(competition)}$\\
\ \ Noise of student observations & 1\% \\
\ \ \# of rounds capturing the utility &  100\\
\ \ \# of random seeds per run &  20\\
\hline
\end{tabular}
\caption{\label{tab:virt_exp} Tabular description of model parameters (Left) and the values taken (Right). Bold signals defaults.}
\label{tab:ve}
\end{table}

\subsection{The Model}

\emph{Attributes for students and schools.} 
Each student and school has an associated vector of attributes. Building on the model proposed by \citet{chen2006school}, the dimensions of these attributes correspond to different evaluation criteria (e.g., level in Math/English, or Science/Arts). The values for the attributes are integer-ratings on a scale from $0$ to $5$. For students, the attributes reflect their current level of knowledge, while for schools it shows their potential to help students. Throughout the experiment, we assume all schools have a maximum potential level; this corresponds to an idealized scenario where schools can help any student achieve the best possible outcome. For students, the attributes are obtained by taking a normally distributed random number and rounding it to the nearest integer on the rating scale. Last, for the presented experiments, we use only one dimension for the attribute vector (e.g., the GPA).
\footnote{In the initial phases of the experiments, we used multiple attribute dimensions. However, once all schools have the maximum potential to help students, the only effect of having multiple attributes is that of changing the mean and standard deviation of the outcome-value distribution. Therefore, we only use one attribute but vary the mean and standard deviation of their distribution.}

\emph{Outcomes.} When a student is assigned to a school, the school decides on the outcome of the interaction. The available outcomes depend on students' and schools' attributes. More precisely, on each attribute, the school can either:
\begin{itemize}[nolistsep]
    \item Put in the \emph{standard effort}: the student will exit the school with a value equal to the minimum between their entry knowledge and the potential of the school;
    \item \emph{Help the student improve}. If the potential of the school is higher than the entry knowledge of the student, the school can choose how much it will help the student improve; the maximum help is the difference between the attributes of the school and of the student.
\end{itemize}
Using the notation introduced before, if a school $y$, with attribute $a_y$, is matched with a student $x$ with attribute $a_x$, then $\mathcal{O}_y(x) = \{a|a\in[\min(a_x, a_y), \max(a_x, a_y)]\}$. For example, if a school with a level of $5$ interacts with a student with knowledge $3$, according to our model, the outcome of the interaction is a student-level of $a\in[3, 5]$. Moreover, the help given by the school to the student is equal to $a-3$.

\emph{Value of outcomes.} 
The value of an outcome is equal to the outcome, i.e., the attribute of the student when exiting the school. In other words, $v_y^o(x) = o$. 

\emph{Cost of outcomes.} 
Depending on the outcome, the school encounters a cost. We assume the cost is the total level of help discounted by a factor of $\alpha$, i.e. $c_y^o(x) = \alpha \cdot \left(o - a_x\right)$. 

\emph{Utility of outcomes.} 
The utility of the school is the value minus the cost, i.e. $u_y^o(x) = o - \alpha \cdot \left(o - a_x\right)$. Note that if $\alpha < 1$, helping the student as much as possible always gives the highest one-step utility for the school and this utility is always positive. To account for the case when there are high integration costs, so the returning side prefers not to receive agents (e.g., for refugee assignment or under-performing student re-assignment), we also allow for negative utility. This is achieved by subtracting the maximum student rating (i.e., the constant $5$) from the utility; we test both scenarios (see Table~\ref{tab:ve}).

\subsection{Preference Formation and Strategies}
\emph{Strategies of schools.} We tested two strategies for the schools. First, we have the \textit{truthful} one, in which schools help students as much as they can. Second, we have a \textit{strategic} interaction. Here, schools distinguish between two categories of students: \textit{cheap} (i.e., students that require less than a threshold, $t_y$, of help from the school $y$), and \textit{expensive} (i.e., students that require more help than that threshold). Under strategic interaction, the school treats the students differently depending on the group they are in: if a student is considered cheap, then the school helps the student as much as possible; otherwise, the school only helps the student to achieve less than the best possible outcome. More precisely, if a student is considered expensive, the school will help the student achieve the best possible outcome minus $l\%$ of the maximum rating. We refer to $l$ as the \textit{level of adversarial attack}. Going back to the previous example, for a school of level $5$, a student of knowledge $3$ considered expensive by the school, and a level of attack of $4\%$, the school will only help the student reach an outcome of $4.8$, instead of $5$.
\footnote{The best threshold depends on the particularities of the system -- especially on the competition and on the sign of the utility per student. Hence, we set the default threshold depending on the parameters. More details on how we do this are in the appendix.}

\emph{Prediction Algorithm.} 
Similar to previous work, we use k-Nearest Neighbour (KNN) to predict academic performance \cite{kabakchieva2010analyzing, asif2014predicting}. More precisely, based on the history of interactions in the most recent rounds (years), the algorithm finds the closest $k$ past students in terms of entry attributes assigned to each school, average their outcome, and use this average as the prediction for the current student.
The prediction is prone to some observation noise which reduces its accuracy. To model it, the predicted outcome varies by $\pm \mathcal{U}(5\cdot p/100)$. We alter the level of noise (see Table \ref{tab:ve}).

\emph{Student preferences.} Students form their preferences based on the predictions and their own observations. We assume the observation of the student is given by the prestige of the school, i.e. the average evaluation score of the outcomes of students who attended the school in the previous year. For each school, the student weights the school as the linear combination between the predicted outcome and their own observation. The importance given to predictions (i.e., \emph{level of trust}) is varied, as shown in Table \ref{tab:ve}. 

\emph{Matching mechanisms.} We implemented $3$ commonly used school choice matching mechanisms, namely Serial Dictatorship (SD), Boston, and Deferred Acceptance (DA)\footnote{See the appendix for a description of each mechanism}.
In practice, these mechanisms differ depending on the ordering of students. For the serial dictatorship mechanism, the students could either be ordered at random (RSD) or, as in, e.g., Mexico City \cite{dustan2017flourish}, by the exam-measured entry level (SD). Similarly, for DA and Boston, the preference of schools could either be given by a random order, or by the true preference of schools. When deployed in the school choice setting, the preferences of schools are usually given by a lottery with some priority ordering (e.g., students with siblings at the same school have a higher priority) \cite{abdulkadiroglu2006changing}. Unless explicitly mentioned otherwise, DA and Boston refer to their respective versions using lotteries.


\subsection{Outcomes Measured in the Experiment}
We run the experiment to capture four key aspects regarding adversarial interaction attacks.

\emph{Incentives to attack}. To see whether schools have an incentive to interact strategically, we compare the utility of one school
\footnote{The values (e.g., utilities) obtained via simulations are in fact the mean of the values obtained by running the simulation setup several times but with different random seeds. When comparing two values we say that their difference is significant if they are more than the sum of their standard deviations apart.}
, $A$, under two scenarios: (a) when all schools, including $A$, interact truthfully and (b) when $A$ interacts strategically while all other schools interact truthfully. If the utility of $A$ increases when it unilaterally interacts strategically, then $A$ has an incentive to deviate from interacting truthfully and adopt such a strategy. 

\emph{The effect of accuracy of and trust in predictions.} 
The impact of predictions depends on two key factors, namely their accuracy and trust. Consequently, system designers usually try to improve on those two metrics \cite{mcnee2006being, dietvorst2018overcoming}. We analyze how such improvements impact the benefits from adversarial interaction attacks.

\emph{Best responses of other schools.}
We use a game-theoretic framework to understand what the response of other schools will be once a school interacts strategically.
To do so, we consider a game with schools as the agents. For computational tractability, we restrict the set of actions to adversarial interaction attacks of levels $0\%, 25\%, 50\%, 75\%$, and $100\%$. The utility for each action is the expected long-term utility according to the simulation. 
With respect to this set of actions, each school has a best response (i.e., an action that gives it the highest utility) given the actions of others. Initially, all schools have an attack level of $0\%$. Then, they take turns finding their best response to the current profile (i.e., they find an action that increases their expected utility the most and significantly). If this process terminates, then the final action profile is a \textit{Nash Equilibrium} (i.e., a choice of action when all schools are playing a best response to the other's actions and, thus, none of them can change their level of attack to achieve a higher expected utility). 

\emph{Student Welfare.} Interacting strategically also affects the welfare of students. Last, we report how this metric, as measured by the average outcome of students, changes as schools adapt their strategies.

\section{Results}
\label{sec_Res}

\subsection{Not all Systems are Interaction-proof}

\begin{figure}[t]
\centering
  \includegraphics[width=1\linewidth]{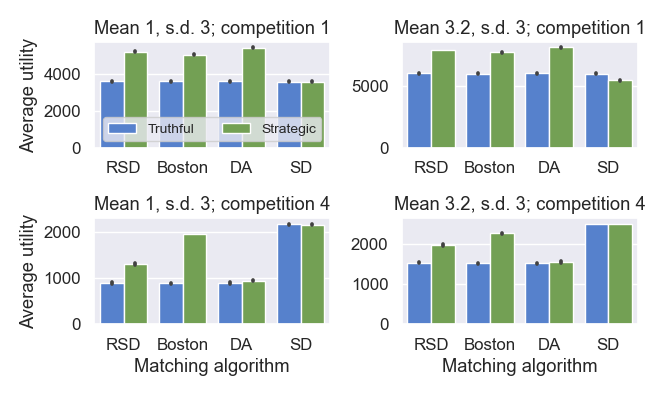}

  \caption{The utility of one school when it interacts truthfully and strategically; the other schools interact truthfully; we also vary the matching mechanism, the distribution of student attributes, the competition level (\# students/place).}
\label{fig:1a}
\end{figure}
As shown in Figure~\ref{fig:1a}, schools usually have an incentive to interact strategically under matching mechanisms that use lotteries.
\footnote{For Figure~\ref{fig:1a}, the student competition level is at least $1$; in this case, setups with positive and negative utility are qualitatively equivalent, so we only included plots for the positive utility setups.}
The utility gain depends on the particularities of the setting matching algorithm and the competition level; for example, when there are as many students as places in schools and DA is used, the utility of the attacking school increases by over $50\%$.
When the preferences of schools are given by the entry knowledge of the student, attacks are no longer beneficial (see SD). This suggests that, when accurate and trusted predictions are available, market designers should consider letting the returning agents express their preferences freely.
In addition, the distribution of the initial levels of students affects the gains obtained through attacks -- schools gain more with the decrease in mean level. 
\footnote{Schools also gain more with the increase in variance. See the appendix for the extended version of Figure~\ref{fig:1a}
.}

\begin{figure}[t]
  \centering
  \includegraphics[width=1\linewidth]{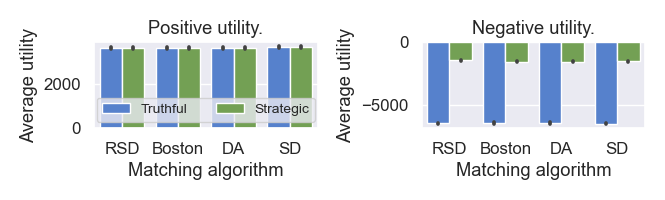}
  \caption{The utility of one school when it interacts truthfully and strategically; all other schools interact truthfully; the competition level is $1/4$; the utility per student is, in turn, positive and negative.}
    \label{fig:1b}
\end{figure}
When there are more available spots in schools than students to fill them, schools do not have an incentive to interact strategically when utilities are positive, but do when utilities are negative (see Figure~\ref{fig:1b}). For positive utilities, schools compete for students; so, even when interacting strategically, the school sets a threshold of $5$ and gives the maximum help to all students. Hence, the strategic and the truthful behavior yield the same average utility. Differently, for negative utilities, schools prefer not to receive students. When the competition level is $1/4$, the school sets a negative threshold and attacks the interactions with all students. Consequently, predictions indicate lower outcomes for students at the school which make most students rank the school last.
\footnote{Although the utilities of schools are computed over a period of $100$ rounds, we point out that schools start having an incentive to interact strategically much faster. In fact, a school has a significant boost in utility by implementing the attack starting from its third round of using it. The plot is included in the appendix.}

\subsection{The Effect of Higher Accuracy and Trust}
\begin{figure}[t]
\centering
  \includegraphics[width=1\linewidth]{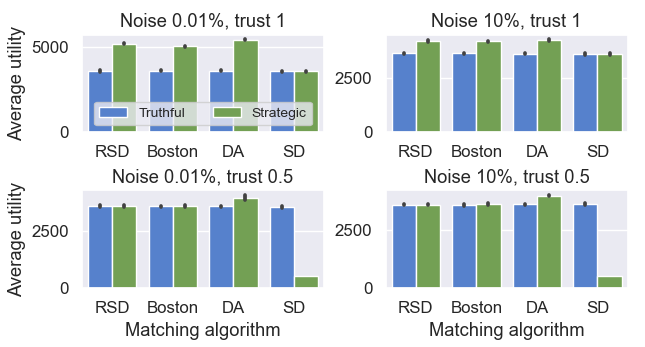}

  \caption{Figures showing the utility of one school when it interacts truthfully and strategically; all other schools interact truthfully. We vary the matching mechanism, trust in predictions, accuracy of predictions.
  }
\label{fig:2}
\end{figure}

Figure~\ref{fig:2} shows that, as predictions become more accurate and trusted, schools gain increasingly more by interacting strategically (under mechanisms using lotteries) and lose less (under SD). In fact, for the RSD and the Boston mechanisms, when predictions have $10\%$ noise and $0.5$ trust levels there is no significant utility gain from attacking. 
This suggests that there are situations in which improving accuracy of and trust in prediction mechanisms could (further) incentivize agents to use adversarial interaction attacks.

\subsection{Responses of Other Schools to Attacks}
\begin{figure}
  \centering
  \includegraphics[width=0.8\linewidth]{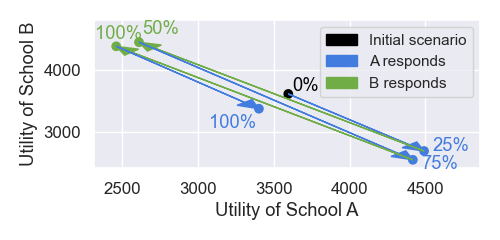}
  \caption{The utilities of schools as they adapt their strategies. Initially, both schools interact truthfully. Each, in turn, changes their strategies to best respond. Arrows indicate the direction of change and annotations the new level of attack.
  }
  \label{fig:best_resp}
\end{figure}

A non-attacking school that competes against schools using adversarial interaction attacks also benefits from implementing such an attack itself.
According to the best response analysis, each school, in turn, adopts an attack of a higher level than the one previously used by the other school. The results for simulating a two-school scenario are shown in Figure~\ref{fig:best_resp}. First, both schools interact truthfully. Next, School A best responds by attacking at a level of $25\%$; this change in action increases the utility of School A and decreases the utility of School B. Similarly, in the next round, School B increases its utility at the expense of the utility of School A by attacking at a higher, $50\%$, level. This continues until both schools attack at the maximum level (i.e., they do not help students considered expensive at all, thus leaving them at their entry level). From this point, neither school benefits from unilaterally changing its action; therefore, this is a Nash Equilibrium. This equilibrium is, however, undesirable for both schools, as it is Pareto dominated by the initial choice of actions; i.e., if both schools interact truthfully, then both of their utilities are higher than when the schools attack at the maximum level.


\begin{figure}
  \centering
  \includegraphics[width=1\linewidth]{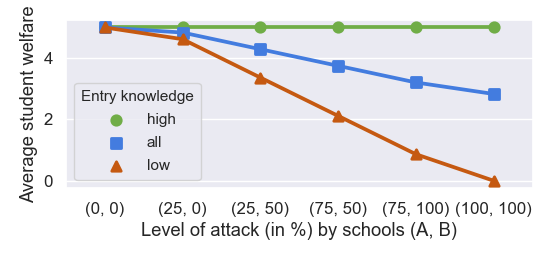}
  \caption{The average welfare of students depending on the levels of attack of schools A and B. Students are grouped based on their entry level of knowledge: all (any level between 0 and 5), low (level 0), high (level between 1 and 4).}
  \label{fig:student welfare}
\end{figure}

\subsection{The Effect of Attacks on Student Welfare}

Figure~\ref{fig:student welfare} shows the average outcome of students depending on the attack levels of each school. As expected, adversarial interaction attacks decrease the welfare of students (as measured by average outcome) proportional to the attack level of schools. In particular, when attacking at a full level, the average outcome of students drops by $43.2\%$. This also increases the disparity between students, as not all are affected equally by the attacks. All loss in welfare is supported by the students targeted by the attacks, i.e., those with a low initial level of knowledge and which are thus considered expensive by the schools. For the chosen distribution of attributes, the targeted students are the ones having a low (i.e., 0) entry level of knowledge. Altogether, Figures~\ref{fig:best_resp} and~\ref{fig:student welfare}  show that the equilibrium with respect to the considered set of actions is non-optimal for both schools and targeted students.


\section{Conclusion}
Predictive models are increasingly used to inform preference-formation in matching markets (MM). However, the robustness of the prediction mechanism under adversarial attacks and of MMs under strategic behavior are usually investigated separately. In the present work, we extend existing models by including the interacting stage in which agents, matched by the MM, interact with each other, thus generating new training data. Doing so makes the feedback loop between the prediction model and the MM explicit and uncovers a new type of strategic behavior: the agents that return to the market in subsequent rounds can deviate from the most profitable interactions in the current round in order to attack the predictions and matchings of future rounds.

Using school choice as an example, we develop an agent-based model to investigate when schools benefit from attacks and what is the effect of attacks on the welfare of students. While we find important differences across varying assumptions about the market, three claims generally hold across those assumptions and have real-world implications. First, the attack we study is \emph{more effective} as predictions get \emph{better} in terms of accuracy (and trust). In alignment with previous work on recommendations \cite{mcnee2006being}, this suggests we should look beyond accuracy when designing prediction mechanisms for such systems. Second, we find that when schools choose to adopt this attack, it has perverse consequences for student utility, in that it both lowers the overall utility \emph{and} increases inequality. These issues reflect how strategic behaviors of social institutions (schools) cause social inequalities, independent of individual potential.
Altogether, this work indicates that both aspects of the matching and prediction mechanisms are key in developing robust systems and sets the framework for a dialog between the ML and MM communities. 
\bibliography{refs}

\section{Appendix}
The Appendix is referenced at multiple points during the main text. Table \ref{tab:footnotes} links the main-text footnotes which point to the Appendix to the related sections within the Appendix. In addition, we also include examples of interaction-proof systems, an explanation of the effects of transitioning from human-based predictions to algorithmic-based predictions, and information of the computing infrastructure for running the experiments.

\begin{table}[t]
\footnotesize
\begin{tabular}{p{0.5 cm} p {7 cm}}
\hline
F\# & Section \\
\hline
1 & Problem formulation - extension\\
2 & Decision problem - assumptions and notation\\
4 & Experiment - multiple dimensions for attributes\\
5 & Experiment - threshold for attacks\\
6 & Matching Mechanisms\\
9 & Results - extended version of Figure 2\\
10 & Results - how fast do schools benefit from attacks\\
\hline
\end{tabular}
\caption{\label{tab:footnotes} Table showing where in the Appendix we are addressing each footnote from the main text.}
\label{tab:footnotes}
\end{table}

\subsection{Matching Mechanisms}
\emph{SD and RSD.} One straightforward example of a matching algorithm is the serial dictatorship mechanism. The students are considered in some order (e.g., in the order of their GPA, or at random), and each student gets allocated to the first school in their ranking that still has available places. Random serial dictatorship (RSD) refers to the version of the mechanism when students are ordered at random. For our experiments, we use SD to refer to the version of the mechanism when students are ordered by their entry level of knowledge. 
\footnote{For SD and RSD, the preferences of schools are not used directly. Instead, the mechanism designer makes an implicit assumption about the preferences of the schools, depending on the student ordering they are using. For example, if students are considered in the order of their GPA, then the implicit assumption is that schools prefer students with higher GPA, while if the ordering is at random, the assumption is that schools are indifferent between which students they are matched with.}



\emph{Boston.} The Boston mechanism proceeds by rounds:
\begin{itemize}
        \item \emph{Round 1.} Students apply to their first choice according to their preference. Schools accept the most preferred applicants (i.e., the students ranked highest), subject to capacity constraints.
        \item \emph{Round k.}  Currently unassigned students apply to their k-th choice (if such a choice exists in their ranking). Schools accept the most preferred applicants from that round, such that they do not exceed their remaining places. 
        \item \emph{End.} The procedure terminates when either (a) there are no more unassigned students, or (b) the unassigned students do not have any school left to apply to.
    \end{itemize}
    
\emph{DA.} The Deferred Acceptance (DA) mechanism is similar to Boston. However, depending on the round, there are some key differences:
    \begin{itemize}
        \item \emph{Round 1.} Schools \textbf{tentatively} accept the most preferred applicants.
        \item \emph{Round k.} Currently unassigned students apply to their most preferred school \textbf{to which they did not apply before}. Schools consider both the applicants form the current round and the tentatively accepted students from before and choose the students they prefer the most. This forms a new set of tentatively accepted students.
    \end{itemize}

\subsection{Problem Formulation - Extension - Human-based Predictions}
Note that our model is agnostic to where the predictions are formed (e.g., in the school choice setting, it is agnostic to whether a guidance counselor  \cite{nathanson2013high} or a statistical model is making the predictions  \cite{wilson2009smartchoice}). Previous work shows that not only are algorithms more accurate than humans at making predictions \cite{dawes1989clinical}, but also, in some settings, they are also more trusted by participants \cite{logg2019algorithm}. Therefore, a transition from human-based predictions to algorithmic-based predictions also leads to an increase in accuracy of and trust in predictions. Paired with our findings from the main text, this means that introducing an, e.g., ML algorithm to make predictions in such settings could increase the gains obtained from using adversarial interaction attacks. 

\begin{figure*}[t]
\centering
  \includegraphics[width=1\linewidth]{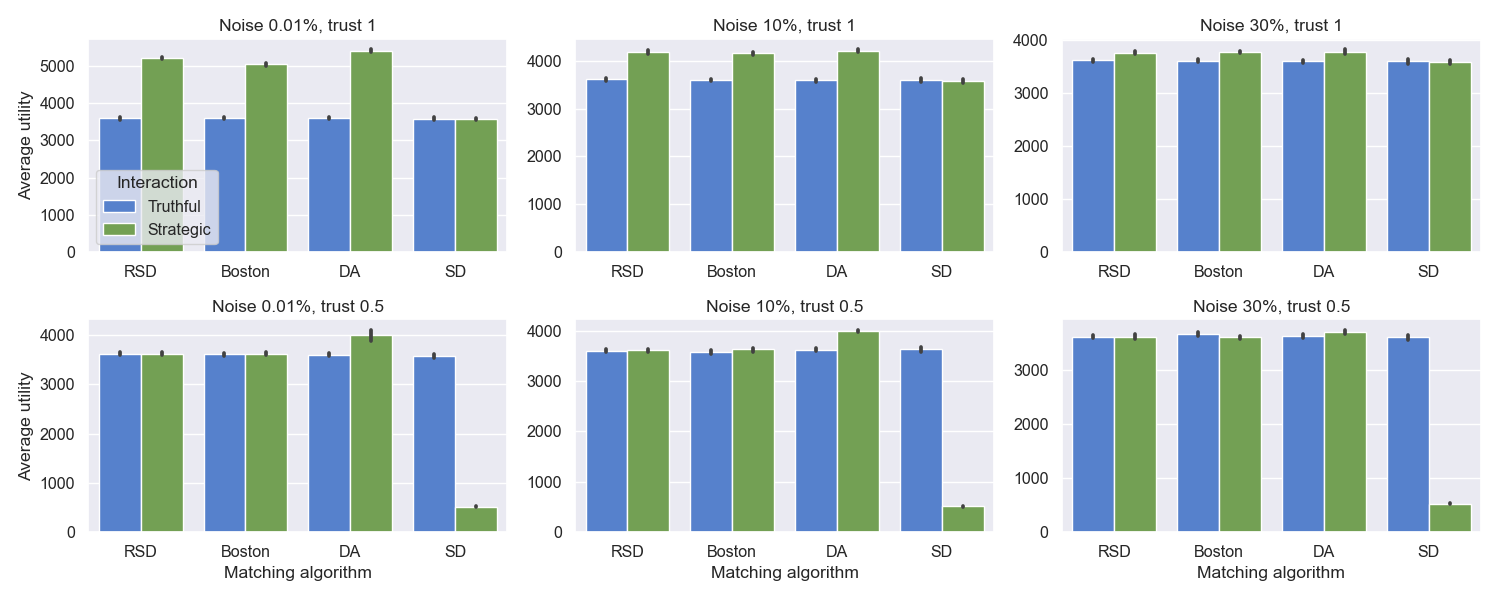}

  \caption{Extended version for Figure 4 from the main text. It shows the utility of one school when it interacts truthfully and strategically; all other schools interact truthfully. We vary the matching mechanism, trust in predictions, accuracy of predictions.
  }
\label{fig:2_ext}
\end{figure*}

Figure \ref{fig:2_ext} extends Figure 4 from the main text. In particular, it shows that when the accuracy and trust levels are low (i.e., $30\%$ accuracy and $0.5$ trust), the non-truthful interaction strategy does not produce any significant gain for any of the tested mechanisms. This underlines the importance of considering adversarial interaction attacks when transitioning from human- to algorithmic-based predictions for preference-formation in such settings.

\subsection{Problem Formulation - Extension - Refugee Assignment}
\begin{figure}[t]
\centering
  \includegraphics[width=1\linewidth]{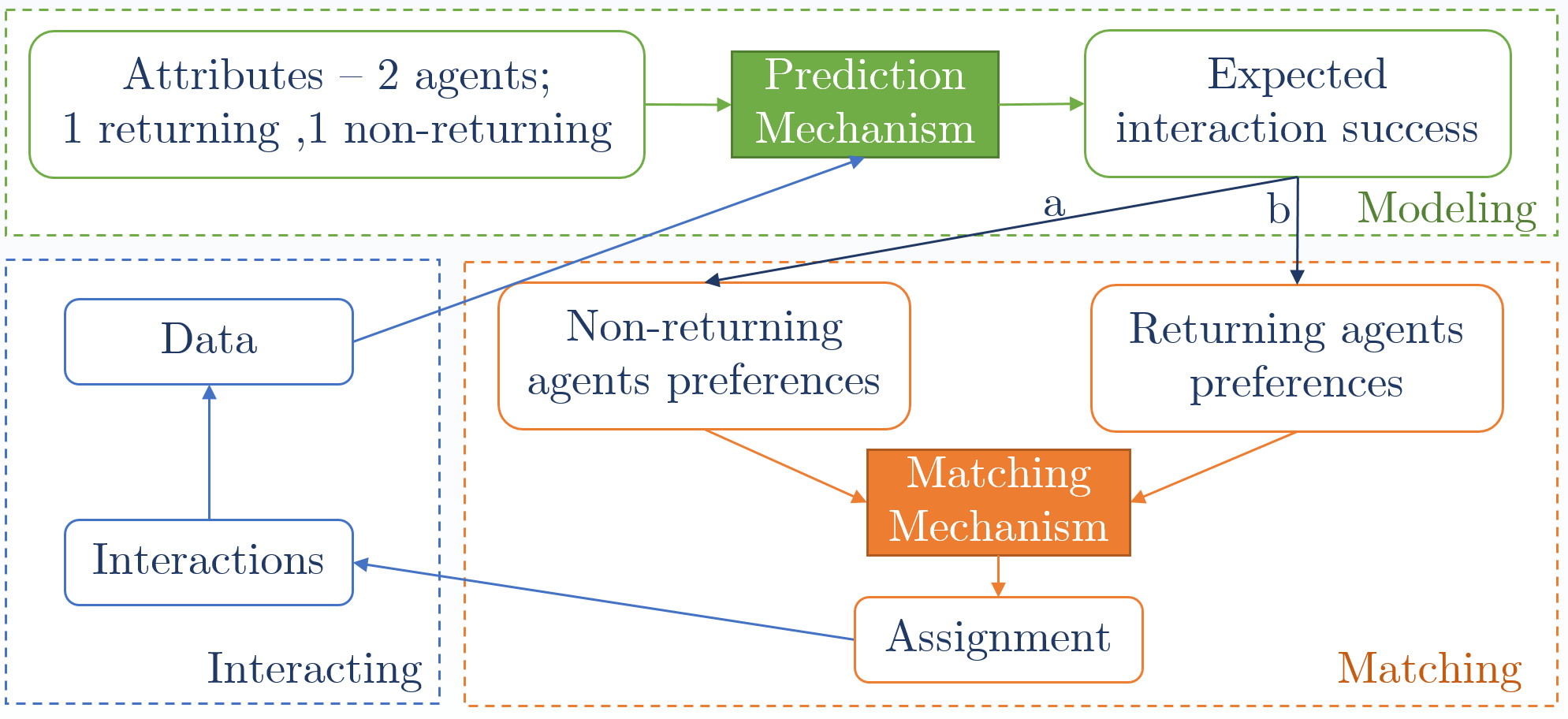}

  \caption{Overview of the generalized three-phased system.}
\label{fig:model_gen}
\end{figure}

As mentioned in the introduction, another application domain using a matching market with prediction-enhanced preference formation is refugee assignment \cite{bansak2018improving, acharya2019combining}. In this case, refugees are the non-returning agents and locations are the returning agents. There is, however, a distinction between how predictions are used in school choice and in refugee assignment. In school choice, predictions help students form their preferences, while in refugee assignment they are used to generate the preferences of locations. Figure \ref{fig:model_gen} shows the extended overview of the system where arrow (a) is used to show predictions that are used to inform the preferences of the non-returning agents (similarly to the school choice setting) and arrow (b) is used to show predictions that are used to inform the preferences of the returning agents (similarly to the refugee assignment setting).

Our problem formulation can be straightforwardly extended to account for this second type of influence too. To do so, we note that the hypothesis, $h$, also induces a ranking over the non-returning side of the market; $h_{\theta_t}(\cdot, y): X\to \mathbb{R}$ gives a weighted preference over $Y$, which we denote by $\succ^{h}_y$. By considering a \textit{prediction integration function} for the returning side too, $h_{\theta_t}^y$, we obtain a way of transforming a (weighted) preference of $y$ into a new preference, $\succ'_y$, using the hypothesis-based ranking for the returning agent, $\succ^{h}_y$. In the case of refugee assignment the integration function simply returns the hypothesis-based ranking, meaning that the locations completely follow the predictions. 

\subsection{Problem Formulation - Extension - Other Application Domains}
In this section we explain how our model extends to three other application domains. First, \citet{paparrizos2011machine} proposed a recommender system (RS) that accurately suggests job transitions based on prior data of employees who changed jobs. In this case, the employees who want to change jobs are the non-returning side and the employers offering jobs are the returning side. The interaction consists of an application/transition assessment (e.g., probation period); such an interaction is successful if the applicant did successfully transfer to the new job position and unsuccessful if the applicant did not. If employers face (a) a cost of assessing candidates, and/or (b) legislation imposing restrictions on which applicants to hire, then they could have an incentive to interact strategically.

Second, also in the context of job recommendations, \citet{liu2016rating} suggested a prediction-based RS to help college students find jobs. In this case, recommendations are based on the similarity between current and past students, and the feedback obtained from past students who attended certain jobs. Similarly to before, students are the non-returning agents and employees the returning agents. The interaction outcome is based on the experience of a student for a certain job.

Third, \citet{kurniadi2019proposed} suggested a RS that suggests courses to students based on performance predictions. For this application domain, students are the non-returning agents and courses the returning agents. The outcome of the interaction is the result of the student (e.g., GPA). Therefore, this application is similar to the school choice setting -- the only difference being that courses replace schools.

\subsection{Problem Formulation - Decision Problem - Assumptions and Notation}
To achieve the succinct problem formulation for the maximum expected discounted utility of a returning agent, we made some simplifying assumptions and notations. First, we assumed $\mu^t$ maps each returning agent to exactly one non-returning agent. Note that this formula can be extended to the general case when each returning agent is assigned a (possibly empty) set of non-returning agents. The only non-trivial step in doing so is to determine the utility of a non-returning agent over a set of interactions (e.g. the utility of a school when interacting with two students could be the sum of the utility in the interaction with each). Second, we did not expand on how the assignment in the next iteration is formed. As a reminder, this is a complex process that depends on a variety of factors, such as the interactions of the other schools, the attributes and original preferences of the students arriving in the next round, the effect of the new history on the RS, the way students integrate the recommendations, and the preferences of the other schools. All these variables produce a (believed) distribution over the possible allocations in the next round. Consequently, the expected value is taken over this distribution.

\subsection{Problem Formulation - Examples of Interaction-proof Systems}
To give two simple examples, a system in which the preference formation is solely based on the attributes of schools (e.g., how far away the school is, what subjects are thought, what are the final examinations) and past data on interactions is not used would be interaction-proof. Similarly, a system that only uses the history of interactions before the decision to introduce the ML-algorithm would also be interaction-proof. This is because, in both of these cases, the future expected utility is constant with respect to the choice of interaction. Thus, the utility from now onward is maximized by choosing the outcome giving the highest utility in the current round.

\begin{figure*}[t]
\centering
	\includegraphics[width=\linewidth]{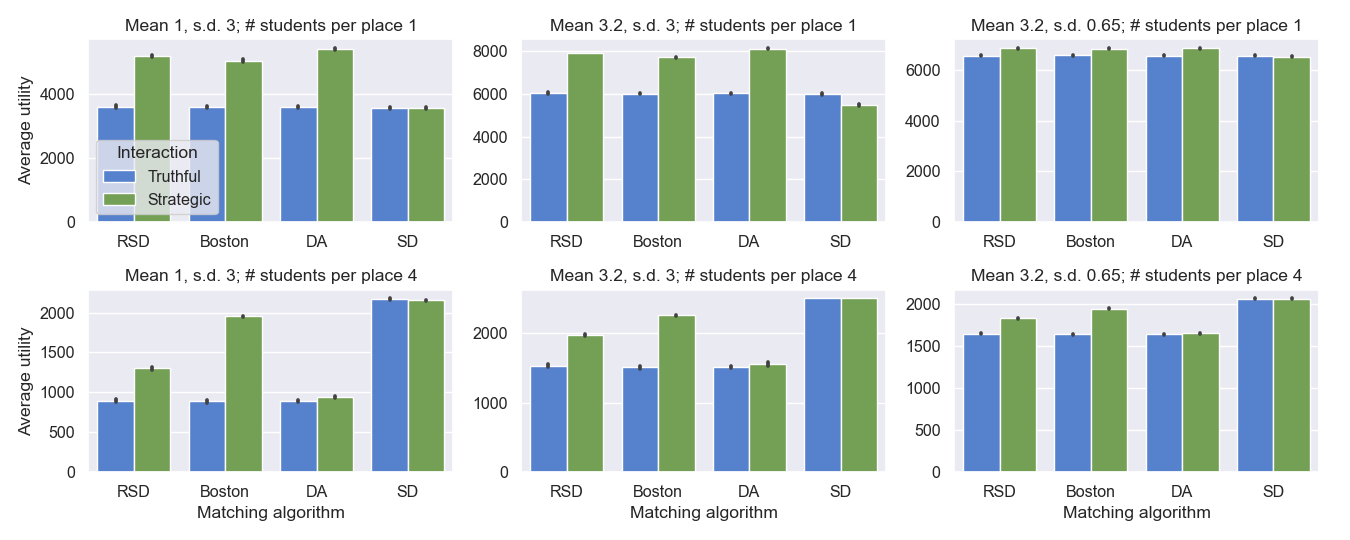}
	\caption{This is an extended version of Figure 2 from the main text. It shows the utility of one school when it interacts truth-fully and strategically; the other schools interact truthfully; we  also  vary  the  matching  mechanism,  the  distribution  of student attributes (i.e., the mean and the standard deviation), the competition level (\# students/place).}
	\label{fig:1a_extended}
\end{figure*}

\subsection{Experiment - Multiple Dimensions for Attributes}
We originally designed the model to account for multiple attribute dimensions (e.g., Math and English scores instead of GPA alone). To do so, we interpret each student attribute as their level of knowledge/expertise in some domain and each school attribute as its potential to help the student improve in that domain. The set of outcomes from an interaction was obtained with the same rules applied component-wise. That is, on each component, a student can achieve an outcome between their original level and the potential of the school. The value of an outcome was the sum of its components.

To give a formal definition of this, assume a school $y$, with attributes $a_y$, is matched with a student $x$ with attributes $a_x$. Then the set of outcomes is $\mathcal{O}_y(x) = \{a|a_i\in[\min((a_x)_i, (a_y)_i), \max((a_x)_i, (a_y)_i)]\}$. As an example let us assume that a school with a level of $3$ for both Math and English interacts with a student with levels $5$ and $1$ respectively. Then, the school has attributes $(3, 3)$, while the student has attributes $(5, 1)$. As a result, according to our model, the outcome of the interaction could be a student level of $(3, x)$, where $x\in[1, 3]$. Moreover, the help given by the school is equal to $x-1$ and the value of the outcome is $3+x$.

Having multiple dimensions of attributes with potentially different distributions of values for each dimension and different functions for computing the values per outcome are all interesting extensions of our work. As a starting point, the publicly available code \footnote{GitHub link: https://github.com/StefaniaI/Predictions-MM.} accounts for the extension mentioned above. In addition, it also allows for measuring the value of an outcome as the minimum of the values per component.

\subsection{Experiment - Choosing the Threshold for Attacks}
For our experiments, when using an adversarial interaction attack, each school $y$ sets a threshold $t_y$. This threshold differentiates between expensive students (i.e., students that require more than $t_y$ help) and cheap students (i.e., students that require at most $t_y$ help). When implementing the attack, the school helps the cheap students as much as possible, and the expensive students less (depending on the level of attack). Its choice of value for the threshold makes an important difference on the strategy. For example, if $t_y$ is $5$ then all students are considered cheap and we obtain the truthful behavior. Differently, if $t_y$ is negative then all students are considered expensive and the school does not help any student achieve their maximum level (for non-zero levels of attack).

This threshold is, thus, a parameter of the strategy in our experiment. Depending on the setting, different thresholds produce better results. Therefore, we choose the threshold depending on the other parameter values. We make this choice based on three factors:
\begin{itemize}
    \item First, schools decide on how many students they want to be matched with. When the utility per student is positive, the school wants as many students as possible, i.e., the minimum between the number of students and the school's capacity. Otherwise, the school only wants students that exceed the capacities of the other schools, subject to their own capacity constraints.
    \item Second, the schools take into account that they will face competition for the students they want. Therefore, when interacting strategically, they aim to get the cheapest $50\%$ of the number of students wanted times the number of schools. 
    \item Third, the schools look at the distribution of attributes and infer the expected number of upcoming students for each level of help. The threshold is obtained using this distribution and the number of students the school aims to get. More precisely, the threshold $t_y$ is the minimum $t$ such that there are at least that number of students considered cheap (i.e., requiring at most $t$ help to achieve the maximum outcome).
\end{itemize}

We illustrate this by an example. Let us assume utilities are negative and there are $3$ schools and $40$ students. If the capacity of each school is $80$ then schools desire $0$ students; differently, if the capacity is $15$ then schools desire $10$ students each. When the capacity is lower there is competition; cumulatively, schools desire the best $30$ students, so, each school will aim to get the best $50\%$ of these students, i.e., the cheapest $15$ students. The remaining ones are targeted by the attack. The threshold is chosen based on the distribution of attributes for the level of knowledge. More precisely, the school $y$ chooses $t_y$ such that the expected number of students coming in the following year and requiring at most $t_y$ help is $15$.

\subsection{Experiment - Computing Infrastructure}
The experiment was designed for Python 3.8.10. For successfully running the simulations one needs the following Python libraries: \textit{numpy}, \textit{pandas}, \textit{scipy}, \textit{csv},  and \textit{copy}. In addition, the visualization functions require \textit{matplotlib} and \textit{seaborn}. We run the simulation on a machine with the following specifications:
\begin{itemize}
    \item OS: Ubuntu 18.04.5 LTS
    \item RAM: 32GB
    \item CPU: Intel® Core™ i7-6700 3.40GHz × 8 cores
    \item GPU: GeForce GTX 1060 6GB/PCIe/SSE2
\end{itemize}
    
\subsection{Results - Extension of Figure 2 (from the main text)}
As mentioned in the main text, the gains obtained from using adversarial interaction attacks change depending on the distribution of the entry knowledge of students: the gains from attacking increase with the decrease in mean and increase in standard deviation. Figure \ref{fig:1a_extended} extends the diagram in the main text by also including the simulation results for a smaller standard deviation value. 

We included here the figure for a mean of $3.2$ and a standard deviation of $0.65$ as this choice of parameters induces a similar distribution to that of the SAT scores of students\footnote{https://reports.collegeboard.org/pdf/2020-total-group-sat-suite-assessments-annual-report.pdf}. However, note that here we assume a one-to-one correspondence between the measured knowledge of a student and the respective value gained by the school. This is not necessarily the case; e.g., a school might not value differently students with scores (out of 800) of 760 and 800, respectively, but might find a larger difference between students with scores of 680 and 720, respectively.

Note that there are important differences depending on other model parameters too. First, depending on the competition level, different mechanisms are more susceptible to interaction attacks. If the number of students is equal to the number of available places, then a school behaving strategically increases its utility the most under DA (by a little over $50\%$). In contrast, if there is a competition of $4$ students per available place at the schools, then Boston is the mechanism that induces the highest increase in utility: in this scenario, the school more than doubles its utility by interacting strategically.

\subsection{Results - How Fast Do Schools Benefit from Attacks}
\begin{figure}[h]
  \centering
  \includegraphics[width=0.9\linewidth]{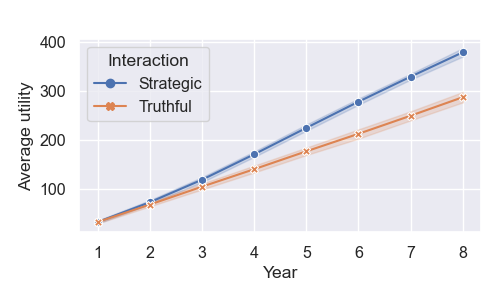}
  \captionof{figure}{Figure showing the utilities with and without strategic interaction depending on the number of years (rounds) elapsed.}
  \label{fig:1c}
\end{figure}

In the main text, the expected utility of one school was computed over a period of $100$ rounds, each round corresponding to a year. However, we want to point out that schools start having an incentive to interact strategically even when considering their utility over a much shorter horizon. In fact, an attacking school gains a significant boost in utility starting from its third year of using it. The cumulative utility of a school with and without strategic interaction over a period of at most $8$ years is shown in Figure~\ref{fig:1c}. 
\footnote{In Figure~\ref{fig:1c}, all parameters are kept at their default values according to Table 1 (from the main text). However, with the increase of the attack level and the decrease of the cost of schools for each unit of help given to a student, $\alpha$, the number of years required to see a significant utility gain by attacking increases. For example, when $\alpha = 0.5$, a $25\%$-level attack only produces a significant gain starting from the fifth year, while a full attack only produces a significant gain starting from the 45th year. Moreover, up to the third and, respectively, 16th year, schools suffer a significant loss in utility by attacking.}

\subsection{Future work}
There are multiple possible extensions for our work. First, we only investigated one type of ML-algorithm; the efficiency of attacks under different algorithms is still unknown. Second, a decisive factor in using adversarial attacks is to be undetectable, which we did not investigate. Introducing good auditing procedures might make some types of adversarial interaction attacks unfeasible. Third, to test the attacks, we used a simple agent-based model. While this was useful for isolating the effects of different parameters and discern their effects, more realistic models, potentially extending to other application domains, will help better understand this type of strategic behavior.  

\end{document}